\documentclass[twoside,11pt]{article}
\usepackage{pgm}
\usepackage{amsmath}
\usepackage{footnote}

\usepackage[utf8]{inputenc}
\inputencoding{utf8}

\usepackage{hyperref,color,soul,booktabs}
\setulcolor{blue}
\newcommand{\BibTeX}{\textsc{B\kern-0.1emi\kern-0.017emb}\kern-0.15em\TeX}

\ShortHeadings{Deep Generalized Convolutional Sum-Product Networks}{Jos van de Wolfshaar, Andrzej Pronobis}

\newcommand{\x}{\ensuremath{\boldsymbol x}}
\newcommand{\scope}[1]{\ensuremath{\mathsf{sc}\left(#1\right)}}
\newcommand{\child}[1]{\ensuremath{\mathsf{ch}\left(#1\right)}}

\begin{document}

\title{Deep Generalized Convolutional Sum-Product Networks}

\author{\Name{Jos van de Wolfshaar\textsuperscript{1,2}} \Email{jos.vandewolfshaar@gmail.com} \and
   \Name{Andrzej Pronobis\textsuperscript{1,3}} \Email{pronobis@cs.washington.edu}\\
   \addr \textsuperscript{1}KTH Royal Institute of Technology, Stockholm, Sweden\hspace{0.5cm} \textsuperscript{2}MessageBird, Amsterdam, The Netherlands\\  \textsuperscript{3}University of Washington, Seattle, WA, USA}

\maketitle

\begin{abstract}
Sum-Product Networks (SPNs) are hierarchical, graphical models that combine benefits of deep learning and probabilistic modeling. SPNs offer unique advantages to applications demanding exact probabilistic inference over high-dimensional, noisy inputs. Yet, compared to convolutional neural nets, they struggle with capturing complex spatial relationships in image data. To alleviate this issue, we introduce Deep Generalized Convolutional Sum-Product Networks (DGC-SPNs), which encode spatial features in a way similar to CNNs, while preserving the validity of the probabilistic SPN model. As opposed to existing SPN-based image representations, DGC-SPNs allow for overlapping convolution patches through a novel parameterization of dilations and strides, resulting in significantly improved feature coverage and feature resolution. DGC-SPNs substantially outperform other SPN architectures across several visual datasets and for both generative and discriminative tasks, including image inpainting and classification. These contributions are reinforced by the first simple, scalable, and GPU-optimized implementation of SPNs, integrated with the widely used Keras/TensorFlow framework. The resulting model is fully probabilistic and versatile, yet efficient and straightforward to apply in practical applications in place of traditional deep nets.

\end{abstract}

\section{Introduction}
    \label{sec:introduction}
    \noindent Sum-Product Networks \citep{poon2011sum} are deep models with unique probabilistic semantics based on a rigorous theoretical framework. They can be seen as both probabilistic graphical models (PGMs) and deep nets, and can be trained using common deep learning techniques (adaptive gradient descent, dropout \citep{peharz2018probabilistic}) as well as those used for PGMs \citep{zhao2016unified,pmlr-v72-rashwan18a}. As opposed to specialized CNNs or GANs~\citep{radford2015unsupervised}, SPNs can perform a wide range of probabilistic inferences efficiently through a single forward and backward pass (including marginal, conditional, joint, MPE), and naturally marginalize out missing data. Whereas CNNs excel at classification, an SPN can perform classification and a variety of generative tasks within a single network. Analogously, GANs, while ideal for sampling, lack probabilistic interpretation and struggle with (or are inefficient for) many generative problems other than sampling. This makes SPNs ideal for domains demanding real-time performance and involving uncertainty and heterogeneous tasks, such as robotics~\citep{pronobis2017iros,zheng2018learning}.

    In this work, we propose Deep Generalized Convolutional Sum-Product Networks (DGC-SPNs) that combine the probabilistic properties of SPNs with the ability to capture spatial relationships in a way similar to convolutional neural networks (CNNs). DGC-SPNs are deep, layered models that exploit the inherent structure of image data and hierarchically capture spatial relations through products and weighted sums with local receptive fields. We introduce a novel parameterization of strides, dilation rates and connectivity for convolutional operations in product layers that makes DGC-SPNs more general than existing approaches to convolutional SPNs \citep{sharir2016tensorial,butz2019deep}. Unlike other architectures, DGC-SPNs employ overlapping patches in product layers, thus avoiding the loss of feature resolution and coverage which enhances representations of spatial relations across the input image, while at the same time preserving the constraints that guarantee validity. We demonstrate that this translates to significant improvements in performance for both generative and discriminative tasks, such as image inpainting and image classification.
    
    Our main contributions are (i) the introduction of a novel convolutional SPN architecture, (ii) a comprehensive range of experiments with SPNs on image data where our DGC-SPNs substantially outperform other SPN models, and (iii) the first layer-centered SPN library \textbf{libspn-keras}\footnote{\url{https://www.github.com/pronobis/libspn-keras}}  built on top of the widely used TensorFlow and Keras frameworks. The library implements the DGC-SPN architecture as well as many other types of SPN models, and provides ease-of-use, flexibility, efficiency, and scalability. Our experiments are easily reproduced using this library.

    
    \section{Background}
    \label{sec:background}
\begin{figure}[!b]
      \centering
      \vspace{-0.5cm}
      \includegraphics[width=0.65\linewidth]{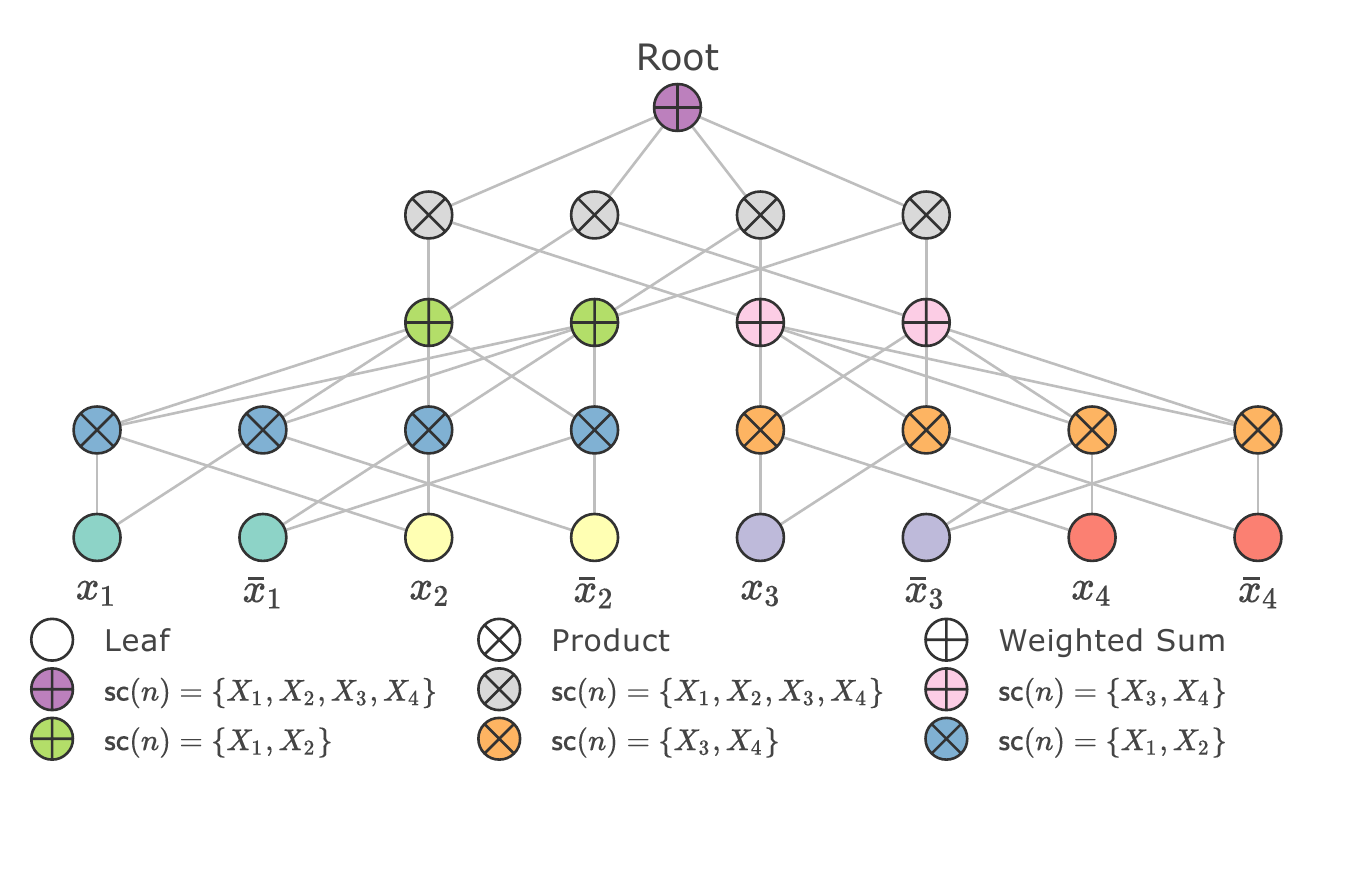}
      \vspace{-0.8cm}
      \caption{An example of an SPN over 4 discrete binary variables ($sc(n)$ denotes the scope of $n$).}
      \label{fig:simplespn}
    \end{figure}

    We begin with a brief introduction of the theoretical background behind SPNs. A Sum-Product Network represents a joint probability distribution over a set of random variables $\boldsymbol X$. An example of a simple SPN is given in Figure \ref{fig:simplespn}. An SPN is a rooted directed acyclic graph where the root node computes the unnormalized probability $S(\boldsymbol x)$ of a distribution at $\boldsymbol x \in \boldsymbol X$. The leaves of an SPN correspond to individual random variables $X_i$. In the discrete case, they are typically represented as Bernoulli variables, while Gaussian distributions are often used for continuous inputs. In between the leaves and the root, an SPN consists of weighted sum and product operations. The weighted sum nodes have non-negative weights and can be interpreted as probabilistic mixture models over subsets of variables, while products compute joint probabilities by multiplying input values and can be seen as features. The output of a sum node $s_j$ is given by $S_{s_j}(\boldsymbol x) = \sum_{i \in \child{s_j}} w_{ji} C_i$ where $C_i$ is the value of the $i$-th child and $\child{s_j}$ is the set of children of $s_j$. The output of a product node $p_j$ is given by $S_{p_j}(\boldsymbol x)=\prod_{i \in \child{p_j}} C_i$. 

    Following \citep{poon2011sum}, we define the following concepts related to SPNs:
    \begin{definition}[Scope] The scope of a node $n$, denoted $\scope{n}$, is the set of variables that are descendants of $n$.  
    \end{definition}
    In other words, the scope of a node is the union of the scopes of its children. Typically, leaf nodes have a \textit{singular} scope containing a single variable $X_i$.
    \begin{definition}[Validity] An SPN is valid if it correctly computes the unnormalized probability for all evidence $\boldsymbol E$ where $\boldsymbol E\subseteq \boldsymbol X$. 
    \end{definition}
    A sufficient set of conditions that ensure validity consists of \textit{completeness} and \textit{decomposability}:
    \begin{definition}[Completeness]
    An SPN is complete \text{if} all children of a sum node have identical scopes.
    \end{definition}
    \begin{definition}[Decomposability]
    An SPN is decomposable \text{if} all children of the same product node have pairwise disjoint scopes.
    \end{definition}
    A unique feature of SPNs is the ability to compute the partition function $Z_S = \sum_{\boldsymbol \x \in \boldsymbol X} S(\boldsymbol x)$ using only a single pass through the network. 
    To this end, all evidence is removed from network's inputs and an upwards pass is computed. For a Bernoulli variable $X_i$, both indicators $x_i$ and $\bar{x}_i$ are set to 1. If all $X_i$ are continuous and represented as multiple Gaussian components per variable, then all of the components corresponding to $X_i$ are set to 1. It has been shown that $Z_{S} = 1$ for a \textit{normalized} SPN, where the weights of each sum node add up to one. Therefore, a normalized SPN computes valid probability with a single upward pass: $S(\boldsymbol x) = P(\boldsymbol x)$  \citep{peharz2015foundations}.  

    \subsection{Parameter Learning for SPNs}
    For discriminative tasks, SPNs can be trained with traditional gradient descent techniques used in deep learning, such as SGD or Adam \citep{Kingma2014AdamAM}. 
    Generative learning is commonly done with expectation maximization (EM). A single EM step requires one forward and one backward pass through the network. As an alternative to vanilla EM, with ‘dense’ non-zero training signals for all sum children in the backward pass, \textit{hard} EM backpropagates sparse signals by selecting only one `winning’ child per sum. For each sum, the child with maximum weighted probability (as computed in the forward pass) is selected as the winning child. The winning child receives a signal of 1 and so its accumulator is incremented by 1 after the corresponding training step, while its siblings receive a signal of 0,  yielding no increments for their accumulators. Yet another alternative is unweighted hard EM which similarly selects one winning child per sum, but relies on unweighted values of the children computed in the forward pass for the selection \citep{kalra2018online}.

    \section{Related Work on Visual Tasks with SPNs}
    \label{sec:related}
    SPNs are expressive and have previously been used in a wide range of domains~\citep{amer2015tpami, zheng2018learning}. Within computer vision, SPNs have been applied to both discriminative and generative problems. In the seminal work~\citep{poon2011sum}, SPNs were used for image inpainting, and assembled following a recursive procedure, where each rectangular image region was split into all possible vertical and horizontal non-overlapping sub-regions. The same architecture was trained with the Extended-Baum Welch (EBW) algorithm to perform image classification \citep{pmlr-v72-rashwan18a}. The non-uniform dimension sizes in this architecture do not lend themselves to GPU-optimized tensorized implementations, limiting their scalabilty. 
    
    The fact that SPNs could benefit from the introduction of convolutions was first recognized in~\citep{hartmann2014friedrich}, where a hybrid model consisting of CNN and SPN layers was used for image classification. However, this yields an architecture that is not fully probabilistic, thereby lacking the ability to perform probabilistic inferences all the way down to the inputs of the model. More closely related to our work are Convolutional Arithmetic Circuits (ConvACs) \citep{sharir2016tensorial}, which can be viewed as convolutional SPNs, and Deep Convolutional Sum-Product Networks \citep{butz2019deep}. Both ConvACs and DCSPNs ensure SPN validity and decomposability of products by using non-overlapping image patches. This effectively reduces feature resolution and fails to capture spatial relations across many regions of the image. In contrast, DGC-SPNs employ a more general approach to convolutions that allow for significantly improved feature coverage and resolution and capture a superset of probability distributions of other convolutional SPNs.

    Apart from spatial SPNs, other works rely on randomly structured SPNs \citep{peharz2018probabilistic}, an SVD-based structure learning algorithm \citep{adel2015learning}, a structure learning algorithm based on clustering of variables \citep{dennis2012learning}, or a structure learning algorithm based on both correlation matrices and clustering \citep{jaini2018prometheus}. Instead, DGC-SPNs take inspiration from CNNs and impose a spatial prior on the structure that corresponds to the inherent spatial relations present in image data. Our experiments show that this translates to gains in performance as DGC-SPNs yield superior results compared to the aforementioned approaches.

    \section{Deep Generalized Convolutional SPNs}
    \label{sec:convspns}
    DGC-SPNs are valid Sum-Product Networks with unique structure inspired by CNNs, consisting of weighted sum and product operations corresponding to local receptive fields. 
    DGC-SPNs are specifically designed to enable efficient utilization of GPU hardware within tensor-based frameworks. DGC-SPN layers are represented as tensors with one dimension for samples in the batch, an arbitrary number of spatial dimensions (2 for images), and one dimension for channels. From~hereon, we omit the batch dimension for simplicity and discuss DGC-SPNs for a single image sample. 
    
\begin{figure*}[!ht]
      \centering
      \includegraphics[width=\linewidth]{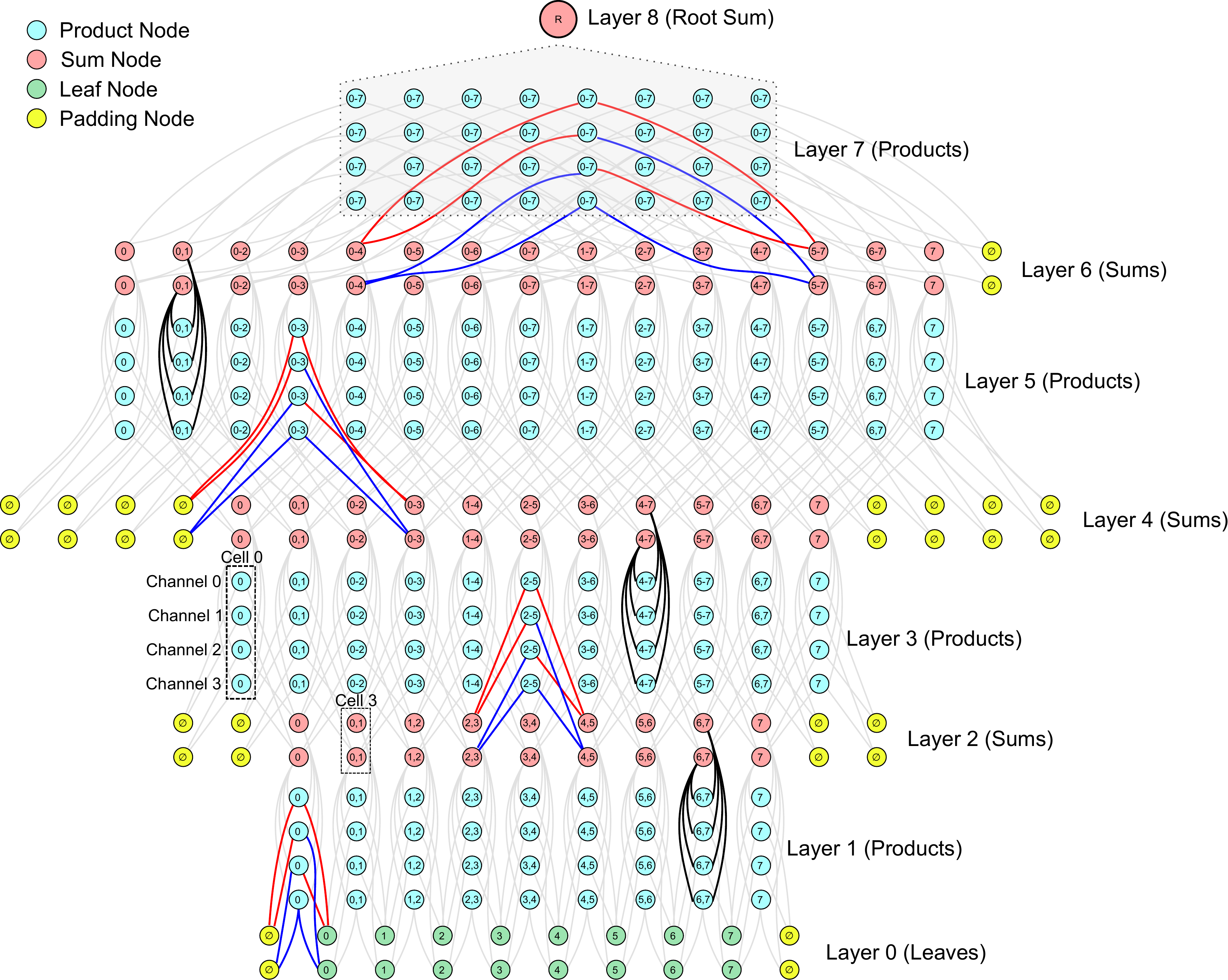}
      \caption{An illustration of a DGC-SPN simplified to 1D. Connections for only one cell per layer are highlighted for readability. Layer 0 contains leaf distributions, where each channel corresponds to an indicator for discrete variables or a distribution component (e.g. Gaussian) for continuous variables. Every product layer doubles the dilation rate, starting at the rate of 1.  The scopes are indicated by the numbers within each node. Padding nodes have a fixed probability of 1 (or 0 in log-space). The nodes of a single cell share the same scope. All children of the root node R have a scope that contains all input variables.}
      \label{fig:dgc-spn-1d}\vspace{-0.2cm}
    \end{figure*}

    The SPN nodes of DGC-SPN (products and sums) are aligned spatially. We refer to all nodes corresponding to a single location $(i, j)$ indexed on the spatial axes as a \textit{cell}. The probabilities computed by the nodes are represented as a tensor $\mathsf X \in \mathbb R^{H\times W \times C}$ for an image with dimensions $W\times H$ and number of channels $C$ (e.g. RGB channels for color images). As shown in Figure~\ref{fig:dgc-spn-1d}, nodes are stacked along the channel axis to form cells, while cells are stacked along the spatial axes to form layers. Sum layers and product layers are stacked in alternating fashion to form a deep network. Sum layers compute mixtures at each cell, resulting in new output channels. Spatial products combine inputs locally to form hierarchical spatial features that exponentially increase their receptive field from the leaf layer to the root.
    
    
    \subsection{Scopes in DGC-SPNs}
    Scopes of nodes within a DGC-SPN have a clear relation to the tensor dimensions mentioned above. If we consider a single cell at the input tensor, we find multiple channels that cover the same variable $X_i$ (e.g. different Gaussian components in case of continuous data or different indicators in case of discrete data). Since channels within a cell cover the same variable, they have the same singleton scope $\{X_i\}$. In other words, scopes \textit{within a cell} of the input tensor are \textit{homogeneous}. In contrast, for image data, different cells at the input correspond to different pixels. If we take any \textit{pair of cells}, we can say that the scopes of these cells are \textit{heterogeneous}. By ensuring that sum layers and product layers preserve within-cell homogeneity, across-cell heterogeneity, completeness and decomposability, we can derive valid convolutional SPN architectures. We now elaborate on how to implement and parameterize such spatial layers.

    \begin{figure}[!b]
      \centering
      \vspace{-0.15cm}
      \includegraphics[width=0.55\linewidth]{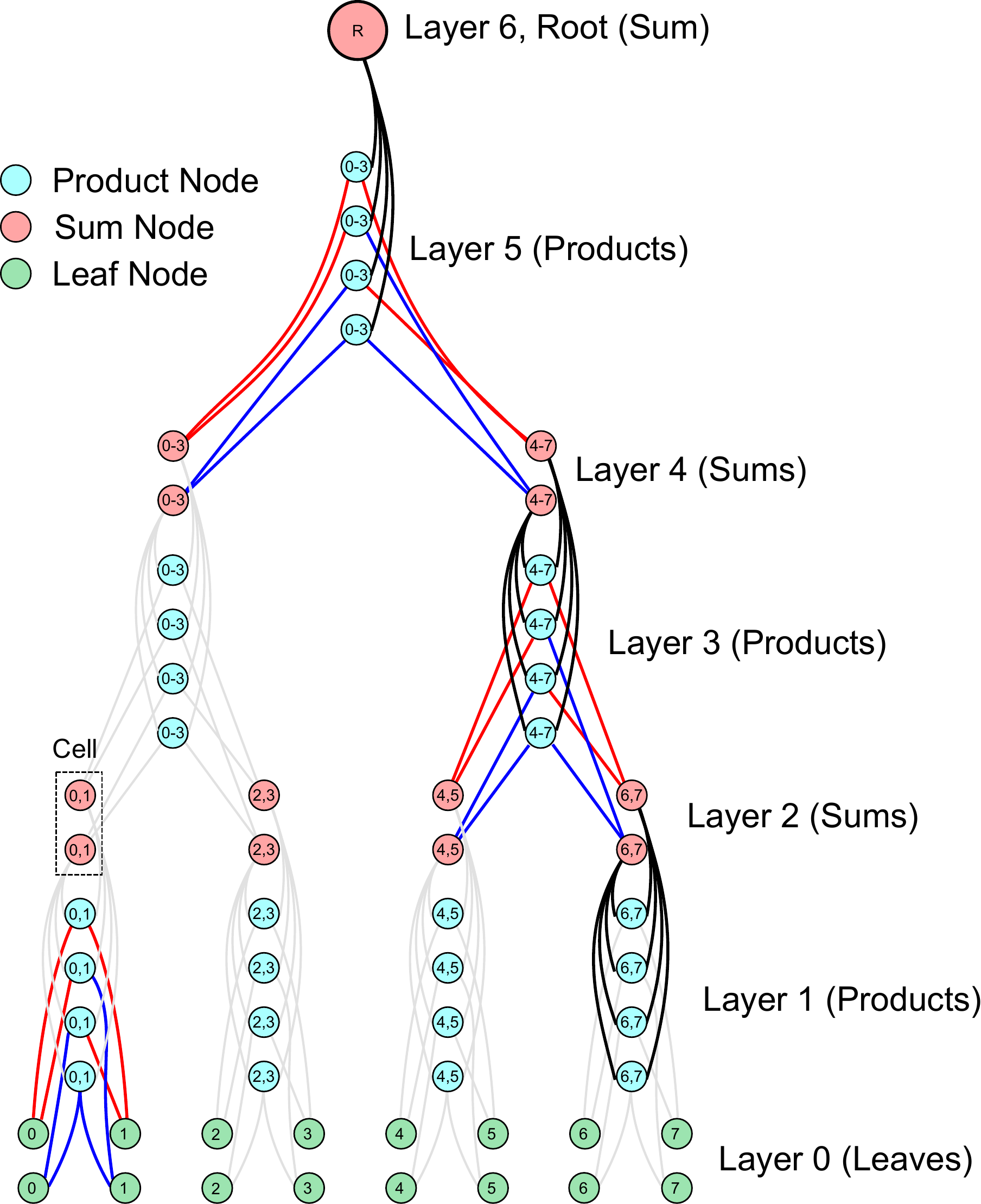}
      \vspace{-0.3cm}
      \caption{An example of a `vanilla' convolutional SPN. As opposed to the more general DGC-SPNs depicted in Figure \ref{fig:dgc-spn-1d}, such architecture does not allow for spatially overlapping patches. Node types are indicated by different colors and some connections are highlighted to improve readability.}
      \label{fig:cspn}
    \end{figure}
    
    \subsection{Spatial Sum Layers} 
    A sum is complete if it has children with \textit{identical scopes}. The within-cell homogeneity and across-cell heterogeneity dictate that at each level, sums should only be connected to a single input cell. Yet, multiple single-cell sums can be added to form an arbitrary amount of output channels. Hence, the spatial layout of the scopes remains unchanged and the validity propagates up the network.
    
    \subsection{Spatial Product Layers}
    \label{sec:spatialproduct}
    Products are decomposable if they are connected to children with pairwise disjoint scopes. As a result, products at each level can have children from at most one channel per cell, but cover two or more input cells. At the input layer, it is trivial to see that neighboring cells are not only heterogeneous, but also pairwise disjoint. Hence, the products on top of an input layer can join scopes by taking small patches of several cells while selecting only one input channel per cell. 
    
    \subsection{Convolutional Log Products}
    \label{clp}
    SPNs are implemented to propagate log-probabilities to avoid
underflow. Hence, the local patches of products become local patches
of sums. In previous works, such local products in log-space were
computed through sum pooling \citep{sharir2016tensorial,butz2019deep},
so that the number of input channels equals the number of output
channels. We propose a more general alternative that implements local
products in log-space through convolutions using kernels with one-hot
weights per cell. One-hot weights are needed so that only one channel
per cell has a coefficient of 1, while all other channels have a zero
coefficient. We refer to such products realized using
convolutions as convolutional log-products (CLP). In Section
\ref{sec:gclp}, we describe an even more general view of CLPs.

    In general, there are $\mathcal C^{t}$ combinations of child nodes
per patch, where $\mathcal C$ is the number of input channels and $t$
is the number of cells under a patch. Consequently, there are at most
$\mathcal C^{t}$ output channels in a CLP layer. To limit that number,
one can take an arbitrary subset of combinations of child nodes. In
this context, sum-pooling can be seen as a special case of CLPs where
the number of output channels equals the number of input channels.
Figure \ref{fig:cspn} displays a convolutional SPN using CLPs
highlighted by the blue and red lines. A red line indicates that the
product is connected to the top channel of the previous layer, while a
blue line indicates that the product is connected to the bottom
channel of the previous layer.


    \subsection{Generalized Convolutional Log Products}
    \label{sec:gclp}
    Existing approaches to convolutional SPNs \citep{sharir2016tensorial,butz2019deep} use non-overlapping patches for their products. Hence, neighboring cells in the output of such layers are not only heterogeneous, but also pairwise disjoint. However, non-overlapping patches require strides larger than 1, thus skipping many combinations of input cells, yielding sub-optimal feature coverage. In fact, this becomes visibly apparent in patch-wise artifacts in image completions \citep{butz2019deep}. 
    
    To overcome these limitations, DGC-SPNs use generalized convolutional log-products (GCLPs). A GCLP is obtained by (i) `full' padding \citep{dumoulin2016guide}, (ii) strides as small as 1, and (iii) exponentially increasing dilation rates. A dilated kernel with a dilation rate of $d$ takes in cells that are $d$ positions apart, leaving gaps of $d-1$ positions. For example, a kernel with $2\times 2$ cells and a dilaton rate of $2\times 2$ (same in both directions), covers a patch of $3\times 3$ cells of the input. The first layer of GCLPs in Figure \ref{fig:dgc-spn-1d} use a dilation rate of 1. We see that the convolutional patches overlap as a consequence of unit strides. Hence, neighboring cells in layer 1 are heterogeneous, but not pairwise disjoint. This forbids us from applying another convolution with a dilation rate of 1. Instead, we exponentially increase the dilation rate to obtain kernels that `skip' the input cells that would otherwise yield non-disjoint children. The exponentially increasing dilation rates yield patches covering pairwise disjoint scopes. Hence, GCLPs exhibit significantly improved feature coverage while preserving decomposability. Full padding is required to ensure uniform dimension sizes in tensors along each axis, allowing the use of GPU-optimized convolution implementations. 

    For readability, the above explanation of DGC-SPNs focuses on 1 spatial dimension and kernel sizes of 2 for all GCLPs. Nevertheless, our architecture generalizes to any number of spatial dimensions and kernel sizes that vary per layer\footnote{In case of varying kernel sizes, the GCLP dilation rates must equal the product of all preceding GCLP kernel sizes.} as shown in our experiments.
    
    \section{Experiments}
    \label{section:experiments}
    We evaluated the generative and discriminative capabilities of DGC-SPNs on two visual tasks: image inpainting and image classification. All experiments were performed using our recently open-sourced \textit{libspn-keras} library. We first describe the experimental setup for both types of experiments.
    
    \subsection{Experimental Setup: Generative Learning}
    
    We assessed generative capabilities of DGC-SPNs for unsupervised image inpainting on a varied collection of image types and datasets. We used images of objects belonging to 101 categories from the Caltech 101 dataset~\citep{li2004learning} and images of faces from the Olivetti dataset~\citep{samaria1994parameterisation}, for which we employed the same $64\times 64$ crops and train and test splits as in \citep{poon2011sum} to ensure a fair comparison. In addition, we used the MNIST written digits dataset~\citep{lecun98gradient} and the Fashion MNIST \citep{xiao2017online} dataset with images of cloth. For these datasets, we used the default train and test splits. All generative experiments were unsupervised, and any class labels in the datasets were ignored. Pixels were normalized sample-wise by subtracting the mean and dividing by the standard deviation for each image.
    
    To build DGC-SPNs, we started with a Gaussian leaf layer followed by a stack of alternating GCLP layers and spatial sum layers. The first GCLP layer computed all possible combinations of products under each patch through one-hot kernels. For the remaining GCLP layers, we used depthwise convolutions (i.e. sum pooling). All GCLP layers considered $2\times 2$ patches with exponentially increasing dilation rates. Between each pair of GCLP layers was a spatial sum layer with 16 output channels. Finally, the top GCLP layer consisted of cells with all variables in their scopes, so that their receptive field covered the full image. This layer was followed by a root sum node.
    
    We trained our generative SPNs with online hard EM using a batch size of 128 for 15 epochs. Although hard EM formally requires MPE inference in the forward pass, following the suggestion in~\citep{poon2011sum}, we used marginal inference instead. For each hard EM iteration, the sum weights were obtained by normalizing the MPE path accumulators with additive smoothing dependent on the number of weights per sum: $w_i = \frac{c_i + \varepsilon}{\sum_j c_j + \varepsilon}$, $\varepsilon=10^{-2} / |\boldsymbol w|$. The Gaussian leaf layer was parameterized by 4 univariate components per pixel. Following \citep{poon2011sum}, for each pixel, the values over all samples in the train set were divided over 4 quantiles. The mean of the $i$-th quantile was used as the value of the mean of the $i$-th Gaussian component at the corresponding pixel, and variances were set to 1. 
    
    For vanilla hard EM, we observed that values of nodes sharing the same sum as a parent start to converge as the depth of the SPN increases. Hence, the impact of sum weights gradually increases, eventually overcoming the impact of the values of the nodes for winning child selection. This forms a self-amplifying chain of signals that only follow the sum child with the maximum weight. Relying on unweighted sum inputs (USI) \citep{kalra2018online} for selection of the winning child mitigates that effect and allows for capturing more complex data patterns during training.

    We performed image inpainting by computing the marginal posterior probability at the Gaussian leaves through partial derivatives \citep{darwich2013differential}. The predicted pixel value is obtained by linearly combining the modes of the leaf components using the marginal posterior probability \citep{poon2011sum}.
    
  \subsection{Experimental Setup: Discriminative Learning}
  
  We assessed the discriminative abilities of DGC-SPNs on the task of image classification. We used the MNIST \citep{lecun98gradient} and Fashion MNIST datasets \citep{xiao2017online}, as these were standard benchmarks for discriminative learning of SPNs in the previous works. We performed sample-wise normalization of pixel values in the same way as for the generative case.

   To build DGC-SPNs for these datasets, we used a Gaussian leaf layer with 32 components per pixel followed by a stack of alternating GCLP and spatial sum layers. For generative experiments, we applied unit strides and exponentially increasing dilation rates with zero padding for all GCLPs. In contrast, here we obtained better results when first using two GCLP layers with non-overlapping strides without zero padding. For the remaining GCLP layers, we applied overlapping strides, exponentially increasing dilations and padding. All GCLP layers used $2\times 2$ patches. We used 64 channels for the first 3 spatial sum layers and 128 channels afterwards.

  We used the Adam optimizer \citep{Kingma2014AdamAM} with its default settings in Keras (learning rate $\alpha=10^{-4}$, decay rates $\beta_1=0.9$, $\beta_2=0.999$). We parameterized the sums with log-space accumulators, denoted $c'_i=\log(c_i)$, so that additional projections onto $\mathbb R^+$ were not necessary after gradient updates \citep{peharz2018probabilistic}. Gaussian means were initialized using equidistant intervals in the range of $[-1.5, 1.5]$ per pixel, and variances were initialized to 1. In contrast to the generative case, the parameters of the Gaussian leaves were adapted after initialization as part of the learning procedure. We used cross-entropy as the loss function.
   
   For regularization, we applied product dropout (PD) \citep{peharz2018probabilistic} by randomly setting product outputs to zero with a rate of $p_{\text{PD}}=0.2$ throughout the entire network. Finally, we applied input dropout (ID) by setting the components of a variable $X_i$ to 1 at random with a rate of $p_{\text{ID}}=0.2$ \citep{peharz2018probabilistic}, as if the dropped out variables were excluded from the evidence. Finally, we used a batch size of 64 and trained our SPNs for 400 epochs.
   
    \subsection{Results}

    Table \ref{resultsuns} shows the results for generative learning and the task of unsupervised image inpainting. We tasked the models with recreating half of an image based on its other half, an extremely difficult problem for a deep architecture. We masked either the left or the bottom half of the images. The performance was assessed based on the mean squared error (MSE)\footnote{MSEs for \citep{butz2019deep} corrected following: \url{github.com/jhonatanoliveira/dcspns/issues/1}.} between predicted pixel values and original images. Pixels were scaled to the range of $[0, 255]$ when computing MSE. 
    
    \begin{table}[t]
      \caption{Results of generative experiments (averaged over 5 runs). We trained DGC-SPNs with and without unweighted sum inputs (USI) and report MSE of predictions for the occluded parts of images.}
      \label{resultsuns}
      \centering
      \begin{tabular}{lllll}
        \hline
        Dataset & Authors & Method & Bottom & Left \\
        \hline
        Olivetti & \citep{butz2019deep} & DCSPN  & $1006$ & $910$ \\
        Olivetti & \citep{poon2011sum} & APVAHRS & $918$ & $942$ \\
        Olivetti & \citep{dennis2012learning} & ClusterVars & $820$ & $814$ \\
        Olivetti & \textit{Ours} & DGC-SPN & $804$ & $847$  \\
        Olivetti  & \textit{Ours} & DGC-SPN + USI & $\mathbf{735}$ & $\mathbf{811}$  \\ \hline
        Caltech 101 & \citep{poon2011sum} & APVAHRS & $3270$ & $3551$ \\
        Caltech 101 & \textit{Ours} & DGC-SPN & $3216$ & $3161$  \\
        Caltech 101 & \textit{Ours}  & DGC-SPN + USI & $\mathbf{2801}$ & $\mathbf{2722}$  \\ \hline 
        MNIST & \textit{Ours} & DGC-SPN & $3767$ & $3102$  \\
        MNIST & \textit{Ours} & DGC-SPN + USI & $\mathbf{2996}$ & $\mathbf{2476}$ \\ 
        \hline
        FMNIST & \textit{Ours} & DGC-SPN & $2870$ & $2268$  \\
        FMNIST & \textit{Ours} & DGC-SPN + USI & $\mathbf{2223}$ & $\mathbf{1637}$ \\ 
        \hline
      \end{tabular}
    \end{table}

 \begin{figure}[t]
      \centering
      \includegraphics[width=.7\linewidth]{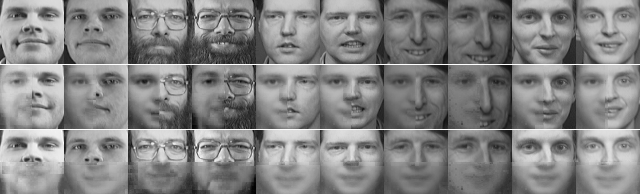}
      \vspace{-0.3cm}
      \caption{Random selection of examples of original images (top row) and inpainting results for left-occluded (middle row) and bottom-occluded (bottom row) test images from the Olivetti dataset.}
      \label{fig:completions}
      \vspace{-0.3cm}
    \end{figure}

    \begin{table*}[!th]
      \caption{Results of discriminative experiments (averaged over 5 runs).}
      \label{resultsmnist}
      \centering
      \smallskip\begin{tabular}{rllllc}
        \hline
        Dataset & Algorithm & Architecture  & Authors & Accuracy  \\
        \hline
        MNIST & EBW & APVAHRS & \citep{pmlr-v72-rashwan18a} & $95.07\%$ \\
        MNIST & DSPN-SVD & Learned  & \citep{adel2015learning} & $97.34\%$ \\
        MNIST & Prometheus & Learned  & \citep{jaini2018prometheus} & $98.37\%$ \\
        MNIST & SGD & CNN + SPN  & \citep{hartmann2014friedrich} & $98.34\%$ \\
        MNIST & Adam & RAT-SPN  & \citep{peharz2018probabilistic} & $98.19\%$ \\
        MNIST & Adam & DGC-SPN & \textit{Ours} & $\mathbf{98.66\%}$ \\
        \hline
        FMNIST & Adam & RAT-SPN  & \citep{peharz2018probabilistic} & $89.52\%$ \\
        FMNIST & Adam & DGC-SPN & \textit{Ours} & $\mathbf{90.74\%}$ \\
        \hline     
    \end{tabular}
    \end{table*}

    
    In all experiments, DGC-SPNs outperformed all other approaches indicating that the generality of our convolutional architecture translates into practical benefits for image data.  Figure~\ref{fig:completions} shows a random selection of completions for images from the Olivetti dataset for one of our experiments.

    The performance gains are apparent even without the use of the unweighted sum inputs heuristic (USI in Table \ref{resultsuns}).
    However, using unweighted sum inputs with hard EM further improves MSEs for all datasets. To the best of our knowledge, our results provide the first quantitative evaluation of this heuristic in the SPN literature. We suggest that its benefits can be attributed to the fact that exponentially decreasing variances cause strong biases in path selection, which can be mitigated by reducing the effect of weights on this process. 
    

   
    The results for the task of image classification are shown in Table \ref{resultsmnist}. In the discriminative learning case, DGC-SPNs again outperformed all other SPN approaches in the literature for both the MNIST and Fashion MNIST datasets.
    
    Several other SPN approaches, including \citep{gens2012discriminative,hartmann2014friedrich}, require sub-SPNs \textit{per class} after which the class-specific SPNs are combined by a single sum node at the root of the SPN. Consequently, these SPN architectures scale poorly with increasing number of classes. In contrast, DGC-SPNs use only a single stack of product and sum layers shared by all classes followed by a top layer of $K$ sums (where $K$ is the number of classes) and a root node resulting in greatly improved scalability.
   
    \section{Conclusions}
    This paper introduced DGC-SPNs, a novel, scalable, deep convolutional architecture for spatial and image data applicable to both generative and discriminative tasks. DGC-SPNs are the most general realization of convolutional SPNs to date, allowing for overlapping convolution patches without breaking validity thanks to a unique parameterization of strides and dilations. This translates to a significant improvement in performance. In our experiments, DGC-SPNs offered state-of-the-art results compared to related SPN architectures on several visual tasks and datasets. 
    
    DGC-SPNs are fully probabilistic, naturally deal with missing inputs, and are capable of efficient joint, marginal, and conditional queries over complex, noisy data. The general design of the model permits its application to a wide range of domains involving spatial data beyond images and two dimensions. By releasing an implementation based on the well-established Keras and TensorFlow frameworks, we hope to encourage further development and application of spatial SPN architectures.

\section{Acknowledgements}
This work was supported by the Swedish Research Council (VR) project 2012-4907 SKAEENet. We would like to thank Avinash Raganath for his indispensable contributions to the LibSPN library and insightful discussions during the time we spent together at KTH. Last but not least, we would like to express our gratitude to Prof. Rajesh P. N. Rao for his unwavering support, encouragement, and invaluable advice.

\bibliography{main}
\end{document}